\documentclass[journal]{IEEEtran}
\ifCLASSINFOpdf
\else
\fi
\usepackage{cite}
\usepackage[utf8]{inputenc} 
\usepackage[T1]{fontenc}    
\usepackage{url}            
\usepackage{booktabs}       
\usepackage{amsfonts}       
\usepackage{nicefrac}       
\usepackage{microtype}      
\usepackage{adjustbox}
\usepackage{amsmath}

\usepackage{graphicx}
\usepackage{comment}
\usepackage{amsmath,amssymb} 
\usepackage{color}
\usepackage{amsmath,graphicx,setspace}
\usepackage{multirow, array, amssymb,booktabs,makecell,xspace,bm}
\usepackage{amssymb}
\usepackage{pifont}
\usepackage{arydshln}
\usepackage{makecell}

\usepackage{amssymb}
\usepackage{pifont}
\usepackage{amsfonts}
\usepackage{arydshln}
\usepackage{dsfont}
\usepackage{multirow}
\usepackage{bbm}
\usepackage{mathtools}

\usepackage{amsfonts}

\newcommand{\etal}{\textit{et al.}}

\DeclareMathOperator{\E}{\mathbb{E}}
\DeclareMathOperator{\F}{\mathcal{F}}
\DeclareMathOperator{\PA}{\mathcal{P}}

\usepackage{url}
\usepackage[ruled,vlined]{algorithm2e}

\usepackage{color,soul}
\sethlcolor{yellow} 
\usepackage{empheq}
\usepackage[many]{tcolorbox}

\tcbset{
  highlight math style={
    colback=yellow,
    arc=0pt,
    outer arc=0pt,
    boxrule=0pt,
    top=2pt,
    bottom=2pt,
    left=2pt,
    right=2pt,
  }
}

\begin{document}
%
\title{Defending Person Detection Against Adversarial Patch Attack by using Universal Defensive Frame}
%
%
%

\author{Youngjoon Yu$^{\dagger}$, Hong Joo Lee$^{\dagger}$, Hakmin Lee, and Yong Man Ro,~\IEEEmembership{Senior Member,~IEEE}
\thanks{This work was supported in part by Center for Applied Research in Artificial Intelligence (CARAI) grant funded by DAPA and ADD (UD190031RD) and in part by BK21 FOUR Program.

Y. Yu, H. J. Lee, H. Lee and Y. M. Ro are with the Image and Video Systems Lab., School of Electrical Engineering, Korea Advanced Institute of Science and Technology (KAIST), 291 Daehak-ro, Yuseong-gu, Daejeon, 34141, Republic of Korea (e-mail: greatday@kaist.ac.kr; dlghdwn008@kaist.ac.kr; zpqlam@kaist.ac.kr; ymro@kaist.ac.kr). Corresponding author: Y. M. Ro (fax: 82-42-350-5494).$^{\dagger}$Both authors are equally contributed to this manuscript.}}

\maketitle

\begin{abstract}
Person detection has attracted great attention in the computer vision area and is an imperative element in human-centric computer vision. Although the predictive performances of person detection networks have been improved dramatically, they are vulnerable to adversarial patch attacks. Changing the pixels in a restricted region can easily fool the person detection network in safety-critical applications such as autonomous driving and security systems. Despite the necessity of countering adversarial patch attacks, very few efforts have been dedicated to defending person detection against adversarial patch attack. In this paper, we propose a novel defense strategy that defends against an adversarial patch attack by optimizing a defensive frame for person detection. The defensive frame alleviates the effect of the adversarial patch while maintaining person detection performance with clean person. The proposed defensive frame in the person detection is generated with a competitive learning algorithm which makes an iterative competition between detection threatening module and detection shielding module in person detection. Comprehensive experimental results demonstrate that the proposed method effectively defends person detection against adversarial patch attacks.
\end{abstract}


%
\IEEEpeerreviewmaketitle

\section{Introduction}

\IEEEPARstart{T}{HE} rise of deep neural networks (DNNs) has achieved substantial success in a wide variety of fields in computer vision \cite{szegedy2015going, wang2017video, khan2018guide, fan2020inf}, including image classification \cite{krizhevsky2012imagenet, he2016deep}, detection \cite{girshick2014rich, girshick2015fast, fan2020camouflaged}, and segmentation \cite{long2015fully, chen2017deeplab}. Despite recent success in various vision tasks, DNNs are highly vulnerable to adversarial attacks \cite{goodfellow2014explaining, zhang2020principal, akhtar2021attack}. Most of these attacks are largely focused on adding small and imperceptible perturbations to all pixels in an image \cite{moosavi2016deepfool, carlini2017towards, madry2018towards, mustafa2019image, li2021toward}. Yet, perturbing all pixels in an image is unrealistic for a real-world attack scenario. Therefore, various methods for physical adversarial attacks have been recently studied due to their practicality \cite{brown2017adversarial, lee2019physical, thys2019fooling, xu2020adversarial, wu2020making, wang2021towards}.

Recent works have demonstrated that the localized adversarial patch attack can fool DNN-based person detectors in the physical world. To be specific, physical adversarial patches can evade person-detection by printing and attaching the adversarial patch to the victim people \cite{thys2019fooling, xu2020adversarial, wu2020making}. To counter the threat of patch attacks in the physical world, some endeavors have been devoted to robust defenses\cite{xiang2020patchguard, chiang2019certified, levine2020randomized, wu2020defending, metzen2020efficient}. There have been also some efforts to adopt existing generic defenses for universal adversarial attacks \cite{borkar2020defending, akhtar2018defense, du2022defending, goel2022fast}. However, those defenses are limited to the image classification domain and do not consider detection details for localized patch attacks. In this paper, we aim to defend a person-detection network because its importance cannot be denied in safety-related applications such as autonomous driving and security systems \cite{rasouli2019autonomous, akhtar2018threat, lin2020graininess, liu2020coupled}.

Compared to the image classification task, protecting person detectors is challenging because of their inherent complexity. In the detection task, a single image can contain multiple people, and the detector should be capable of handling a list of multiple bounding box coordinates and class labels. Furthermore, unlike common universal adversarial attacks, the localized patch-based attack has been widely known as highly dependent on the size of the receptive field \cite{xiang2020patchguard, yu2021defending}. Thus, naively applying defenses from the image classification does not guarantee the end-to-end robustness of the detection against localized patch attacks. Prior works \cite{chiang2019certified, levine2020randomized} required separate forward passes for possible patch locations and heavy computational costs. Therefore, only a few defensive methods are conducted on detection tasks to achieve empirical robustness against adversarial patch attacks~\cite{ji2021adversarial, saha2020role, xiang2021detectorguard}.


Saha \etal~\cite{saha2020role} investigated the role of spatial-context information for adversarially robust detection. They proposed a grad-defense method to limit the usage of contextual information in general detection tasks. But, their defense is limited to the fixed setting of an adversarial patch at the image corner. Thus, the benefit of their approach is not applicable to counter state-of-the-art adversarial patch attacks because recent adversarial patches are placed on the person directly~\cite{thys2019fooling, xu2020adversarial,wu2020making}. In addition, Ji \etal~\cite{ji2021adversarial} proposed an adversarial patch detection network that could defend against adversarial patch attacks. But, their work is based on adversarial training~\cite{madry2018towards}, and thus it requires several rounds to re-train the backbone model with large datasets. Xiang \etal~\cite{xiang2021detectorguard} also suggested a localized adversarial patch detection network and achieved certified robustness. According to their paper, when an attack patch is detected, a model excludes the sample from making predictions (abstention); however, it is not a desirable solution for adversarially robust detection \cite{xiang2021detectorguard}.


\begin{figure*}[!t]
	\centering
	    \includegraphics[width=0.95\linewidth]{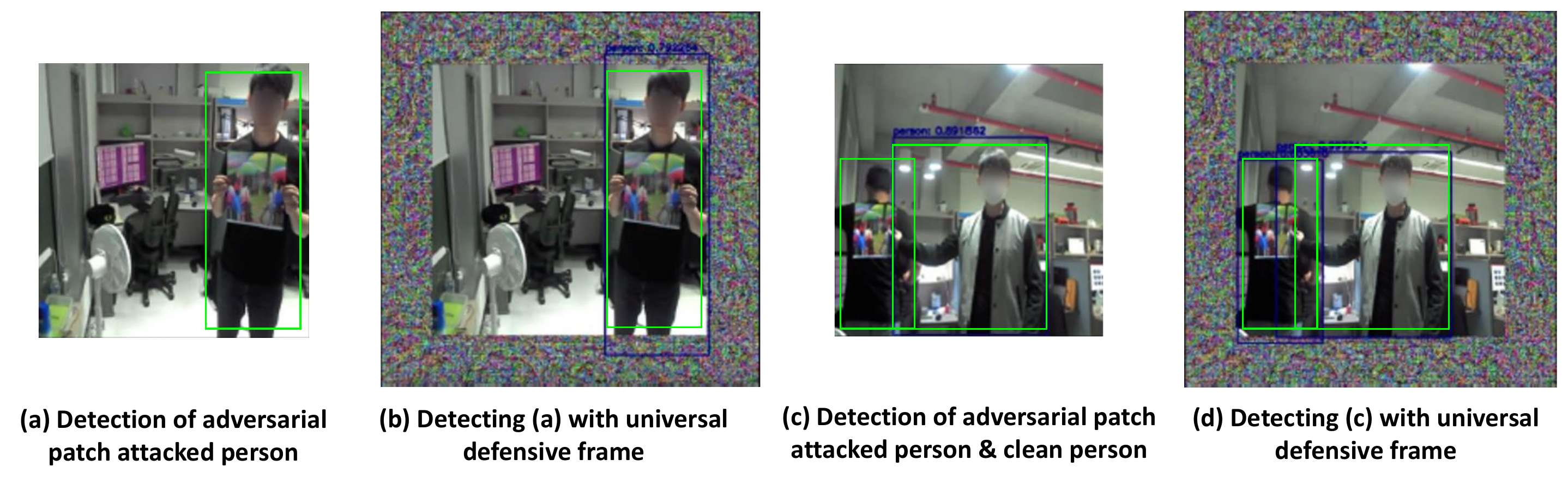}
	\caption{Person detection results of the proposed universal defensive frame (UDF). Ground truth is depicted as green line and the prediction is depicted as blue line. (a) an adversarial patch \cite{thys2019fooling} attacked person evades person detection; (b) the adversarial patch attacked person in (a) can be detected with UDF; (c) the detection of the adversarial patch attacked person and clean person (d) the adversarial patch attacked person in (c) can be detected with UDF. A demo video can be found at: https://youtu.be/hAXZWv95EPE}
	\label{fig:1}
\end{figure*}


From the counter-intuitive perspective, the aforementioned challenges and difficulties of existing countermeasures reflect the inherent strength of localized patch-based attacks. Thus, we do not have attention to marginally improving previous countermeasures, but we rather focus on exploiting a powerful capability of a localized pattern as a defense. The basic intuition behind this is the possible existence of a localized pattern that could invalidate the effect of an adversarial patch. By introducing this kind of defensive pattern, we focus on tackling the following problems to defend the person detection against adversarial patch attacks. First, most defense methods require large computational costs. In other words, to defend against the adversarial patch attack, previous methods re-train the detection network with large datasets \cite{saha2020role, ji2021adversarial} or an additional network for robust inference \cite{ji2021adversarial, xiang2021detectorguard}. Second, there is a performance gap between clean performance and adversarial robustness \cite{zhang2019theoretically}. 

In this paper, we propose a novel person detection defense approach against adversarial patch attacks. We use a defensive local pattern to defend person detection, which is called a universal defensive frame (UDF). Our method optimizes localized patterns that belong to a frame. Figure \ref{fig:1} shows demo video results of the proposed UDF: By applying UDF to the outside edges of images, a person detection network can detect persons regardless of the adversarially-patched image. To defend against the adversarial patch attack, UDF has three properties. First, UDF suppresses the effect of the adversarial patch. Second, UDF maintains the original person detection (detection with clean images). Third, UDF is effective regardless of input images (image-agnostic). To satisfy the aforementioned properties, we devise a novel UDF optimization algorithm for defending person detection. In the proposed algorithm, we create sub-image sets consisting of \textit{M} images sampled from the data distribution $\mu$. The frame-shaped defensive pattern is then optimized to reduce the difference between the output of \textit{M} adversarial images and the output of \textit{M} clean images. By repeating this optimization process, we can coin the defensive frame \textit{universal}, as it represents a fixed image-agnostic pattern that defends most images sampled from the data distribution $\mu$. The optimization is repeated over the entire sub-image set between two opposing purposes competitively: maintaining the original person detection and misleading the person detection. Toward this end, the optimization process generates UDF robust enough to against the adversarial patch while also maintaining image-agnostic detection performance. Furthermore, since we optimize a small and restricted region in the frame, there is no large additional computation cost and UDF can be applicable to any weight-fixed person detector. In other words, we do not have to change the person detection model. 
To conclude the introduction, we outline the major contributions of this work:

\begin{itemize}
    \item We propose a novel defensive approach in person detection, which can suppress the effects of adversarial patch attacks. We devise a novel person detection defense against adversarial patch attacks with a universal defensive frame.
    
    \item To generate the universal defensive frame for person detection, the adversarial patch and the defensive frame iteratively compete with each other in the optimization process of the universal defensive frame.
    
    \item The proposed defense method for person detection does not require large computation costs and re-training the weight-fixed detector model. Rather, it maintains the detection performance of clean images.
    
    \item We perform comprehensive experiments to validate that the proposed method is useful in the defense against adversarial patch attacks in person detection.
    
\end{itemize}

\section{Related Works} \label{sec:2}
\subsection{Adversarial Patch Attack for Person Detector}
Recently, many physical adversarial patch methods that could affect real-world applications have been studied. Lee \etal~\cite{lee2019physical} generated a physical adversarial patch that could suppress all the detected objects in the image. By considering the physical factors (e.g., illumination, angle, scale), they produced a physical patch that works in the real world.
Thys \etal~\cite{thys2019fooling} demonstrated that a physical adversarial patch can hide a person. They generated the physical adversarial patch in real-world person detection by reducing three loss functions: objectness score loss, total variation loss, and non-printability score loss. The objectness score loss reduces the objectness score in the image. The total variation loss enhances the smoothness of the adversarial patch. Finally, the non-printability score loss explains how well the colors in the patch can be physically represented by a printer. By reducing this loss, the adversarial patch can be printed in the physical world. Unlike previous adversarial patches placed on the background, their patches are more powerful because they are placed on the person. Xu \etal~ \cite{xu2020adversarial} produced ‘Adversarial T-shirts’ where the adversarial patch is printed on a T-shirt. They exploit thin-plate spline (TPS) transforms so that, their patches are effective on non-rigid objects (e.g., T-shirts). Wu \etal~ \cite{wu2020making} investigated the transferability of physical patch attacks in person detection. They produced a physical adversarial patch that can evade multiple person detectors with improved transferability. To improve the transferability, they optimize the adversarial patch with ensemble detection models.

\subsection{Defending against Adversarial Patch Attack}
Some defense methods have been devoted to defending against an adversarial patch attack for image classification \cite{rao2020adversarial,chiang2019certified, mccoyd2020minority,levine2020randomized,xiang2020patchguard}. McCoyd \etal~\cite{mccoyd2020minority} proposed the Minority Report defense. It predicts the location of the non-ideal prediction as the patch position while masking the pixels of the input image. Xiang \mbox{\etal~\cite{xiang2020patchguard}} proposed a certified defense method named PatchGuard. It aggregates all classification logit predictions while masking the spatial pixels of the feature, and thus defends against patch attacks by excluding abnormal scores. Although these approaches have shown efficacy against patch attacks, they hardly transferred to detection because their algorithm heavily depends on the classification logit, not on the spatial location information. Furthermore, most of these methods require multiple forward passes for all possible patch locations and are thus computationally expensive.

Other defenses have focused on defending against an adversarial patch attack in detection tasks. Toward this end, Saha \etal~\cite{saha2020role} exploited spatial-context information between objects. To utilize this spatial-context information, they overlay an out-of-context patch on an image. Then, they train the detection network with out-of-context patch images. However, this requires additional training datasets and re-training the detection network from scratch. Ji \etal~ \cite{ji2021adversarial} proposed a plug-in defense component for the YOLO detection model to defend against adversarial patch attacks. It avoids false person detection by training the detection model to detect adversarial patches. Chiang \etal~\cite{chiang2021adversarial} proposed a strategy that defends against adversarial patch attacks by detecting and removing the adversarial patch with an extra detection model called the segmentor. The segmentor segments the region that adversarial patch is located and removes that region. Most of these defense methods can defend against adversarial patch attacks. But, these methods need to re-train the detection network, which requires large computational costs and training time. They also require extra heavy modules for inference and an additional dataset. 

Recently, some studies are utilized in both classification and detection tasks against the adversarial patch. Naseer \etal~\cite{naseer2019local} suggested local gradient smoothing (LGS) to locate and erase abnormally high-frequency areas in the input image. They regarded this area as the adversarial patched area. Yu \etal~ \cite{yu2021defending} suggested the feature norm clipping (FNC) defense to adaptively clip large norm deep features induced by universal adversarial patches. But their defense performance 
relies too much on the threshold selection. Here, we propose a novel defense strategy that optimizes a defensive frame pattern. This frame pattern obviates re-training the model and avoids large computational costs. It is applicable to any pre-trained weight-fixed detector.

\section{Problem Definition} \label{sec:3.1}
\subsection{Adversarial Patch Attack in Person Detection} The objective of the adversarial patch attacks in a person detection is to make detectors miss the detection of a person class at the inference time. DNN-based person detection networks can be categorized into two types: two-stage framework (e.g., Fast(er) R-CNN \cite{girshick2015fast, ren2015faster}, Mask R-CNN \cite{he2017mask}, etc.), and one-stage framework (e.g., YOLO \cite{redmon2017yolo9000}, SSD \cite{liu2016ssd}, etc.). The two-stage framework consists of a region proposal network (RPN) and a head network. The RPN produces an objectness score that identifies potential bounding boxes while the head network classifies the contents of these bounding boxes and regresses the coordinates. The one-stage frameworks detect objects at once. The output of a one-stage framework has the $w \times h$ size of a feature map, and each pixel contains the locations of the bounding boxes, classes, and objectness scores. \cite{thys2019fooling} demonstrated that reducing the objectness score effectively fools the person detection. In this way, the adversarial patch attack reduces the number of true positives. Therefore, the adversarial patch can be optimized by minimizing the following loss function,

\begin{equation}
\label{eqn:1}
\mathcal{L}_{adv}(x,p)={\E_{x}}[\max({\textbf{S}({\PA}\left(x,p\right)), 0)]}.
\end{equation}

\noindent Here, $x$ is the input image, $\PA(\cdot)$ places an adversarial patch $p$ on persons with transformations such as rotating and resizing \cite{thys2019fooling}, and $\textbf{S}(\cdot)$ denotes the output vector of objectness scores. Specifically, $\textbf{S}(\cdot)$ can be the output of the RPN network in the two-stage detection network or the output feature map's objectness score channels in the one-stage detection network. By optimizing the above loss function, an adversarial patch can be generated to fool the person detection.

\begin{figure*}[!t]
	\centering
	    \includegraphics[width=0.95\linewidth]{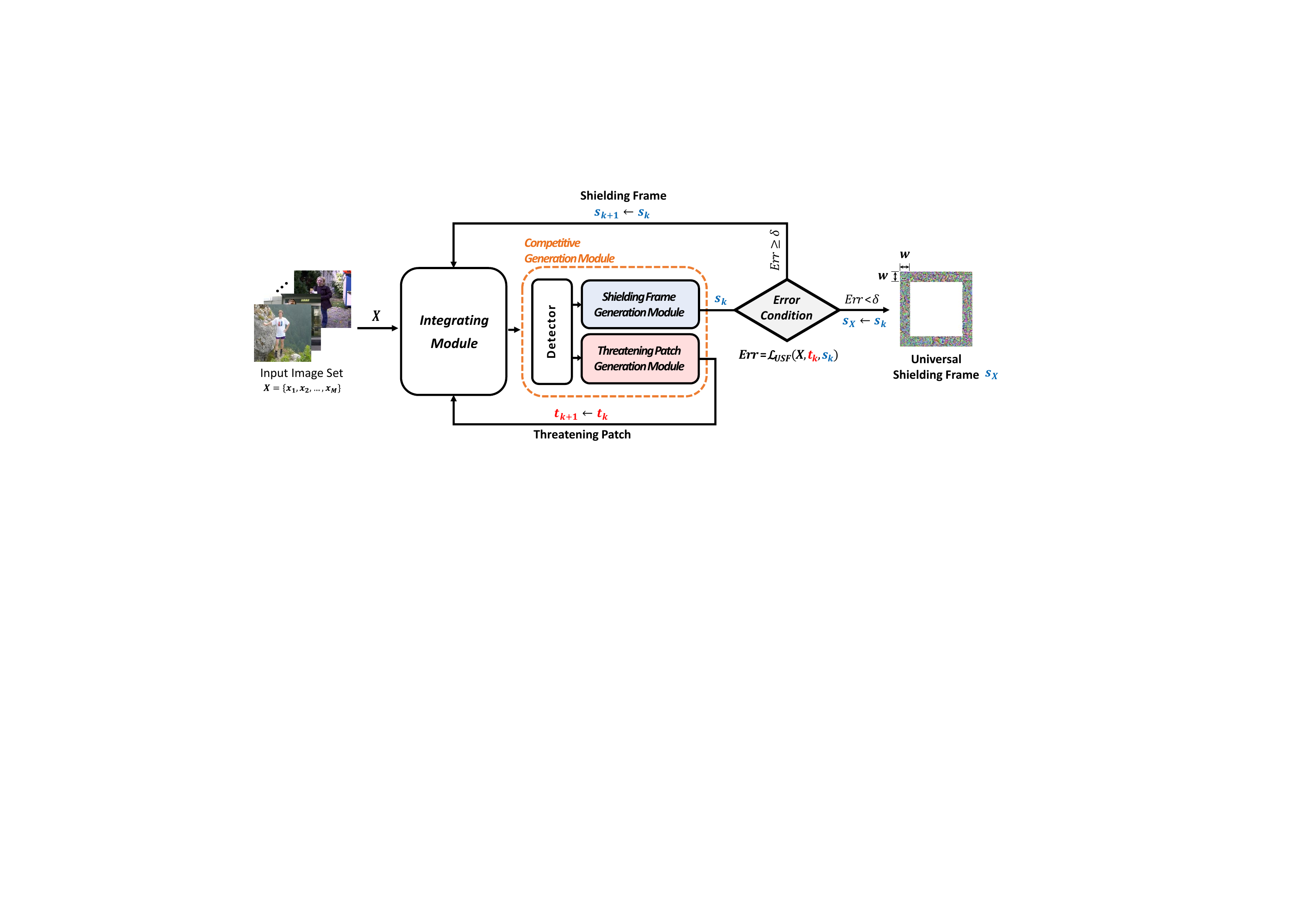}
	\caption{Overview of the proposed framework to generate the universal shielding frame $s$ with frame width \textit{w}. The framework consists of an integrating module, the competitive generation module, and the error condition check. The integrating module combines input image set $X$, shielding frame $s_k$, and threatening patch $t_k$. The competitive generation module (orange dotted box) takes the integrated image set as an input and iteratively returns the shielding frame $s_k$ and threatening patch $t_k$. The error condition modulates the termination of iteration by the switching path according to the error rate threshold $\delta$. After optimizing for all sub-image sets in the entire dataset, we used the final optimized universal shielding frame $s^*$ as universal defensive frame $d$ in the inference stage. $\mathcal{L}_{USF}$ is the loss function to optimize the universal shielding frame.}
	\label{fig:2}
\end{figure*}

\subsection{Defensive Frame in Person Detection}
Securing person detectors from the aforementioned adversarial patch attack has been challenging due to the complexity of the detection task. Attackers use a localized patch for physical world attack \cite{thys2019fooling,wu2020making,xu2020adversarial} that makes the detection network miss person detection. To defend person detectors, recovering the number of true positives is more important than reducing the number of false positives in real-world applications~\cite{saha2020role}. For example, in autonomous vehicle applications, not detecting a person is more dangerous than detecting several wrong persons. In this paper, we focus on a defensive pattern that mitigates the effect of adversarial patch attacks on true positives in person detection. We devise a new defensive pattern for person detection which is against the adversarial patch attack. The main intuition of the existence of the defensive frame stems from the phenomenon of adversarial patches. Like the adversarial patch, if we can mislead the original prediction by manipulating the input, then this phenomenon could actually be turned upside down to alleviate the effectiveness of adversarial patches in the input domain. Since the adversarial patch attack in person detection has become a serious threat in the field of security and surveillance \cite{akhtar2021advances, arnold2019survey, zhang2018pedestrian}, we chose the target defending class as a person. We make the person detector adversarially robust by adding special types of defensive patterns to input images so that a standard detector can maintain the original person detection. To achieve the objective, we obtain the defensive pattern $d$ by minimizing the following loss:

\begin{equation}
\label{eqn:2}
\mathcal{L}_{def}(x,p,d) = {\parallel{f_\theta}(A_{def}(\PA(x,p),d))-f_{\theta}(x)\parallel_r},
\end{equation}

\noindent where $\parallel \cdot \parallel_r$ denotes the $r$-norm distance, and the applying function $A_{def}(\cdot)$ adds a defensive pattern $d$. Term $\theta$ denotes the parameter of the detection network. $f_{\theta}{(\cdot)}$ refers to the detection output which is the combined vector including the classification score, the objectness score, and the bounding box location. Equation (\ref{eqn:2}) optimizes the defensive pattern $d$ to minimize the difference between the prediction of ${f_\theta}(A_{def}(\PA(x,p),d))$ and $f_{\theta}{(x)}$. To generate the best-performing defensive pattern which is well suited to the input image, in this paper, we devise the frame-shaped defensive pattern with the following three properties.

\subsubsection{Maintaining detection performance for original content} For the defender, it is normally impossible to know in advance whether the adversary has attacked the input image. It would be enough for attackers to make changes that the defenders may not notice. But, from the perspective of defenders, the best option is to keep the original prediction without modifying the original content of the input before and after the defense. This is because we do not want to hurt the detection performance of any clean image. To this end, we create a new defensive frame for mitigating adversarial patch attacks, which is added in a border to surround the original input image.

\subsubsection{Locally invariant performance of defensive frame} Attackers in the physical adversarial attack settings can arbitrarily change the location of the adversarial patch according to the target person. The defensive effect of our proposed method should not be significantly changed according to the location of adversarial patches and the location of the defensive frame itself. The natural properties of the defensive frame can meet this location-invariant defensive performance.

\subsubsection{Adjusting the defense level in person detection}
In general, standard adversarial attacks that add epsilon-bounded perturbation to the whole image can change the magnitude of the attack level by controlling the epsilon norm of perturbation. This method is not necessarily suitable for adversarial patches because the patches are not norm-constrained. In other words, the patch attack already has perceivable and sufficiently large perturbations in a locally limited dimension. Thus, it is not easy to change the magnitude of the attack level in a simple way. The defensive frame has the same issue with changing the magnitude of the level of defense. In this paper, we can adjust the defense level by simply changing the frame width of the defensive frame.

\begin{figure*}[!t]
	\centering
	    \includegraphics[width=1.0\linewidth]{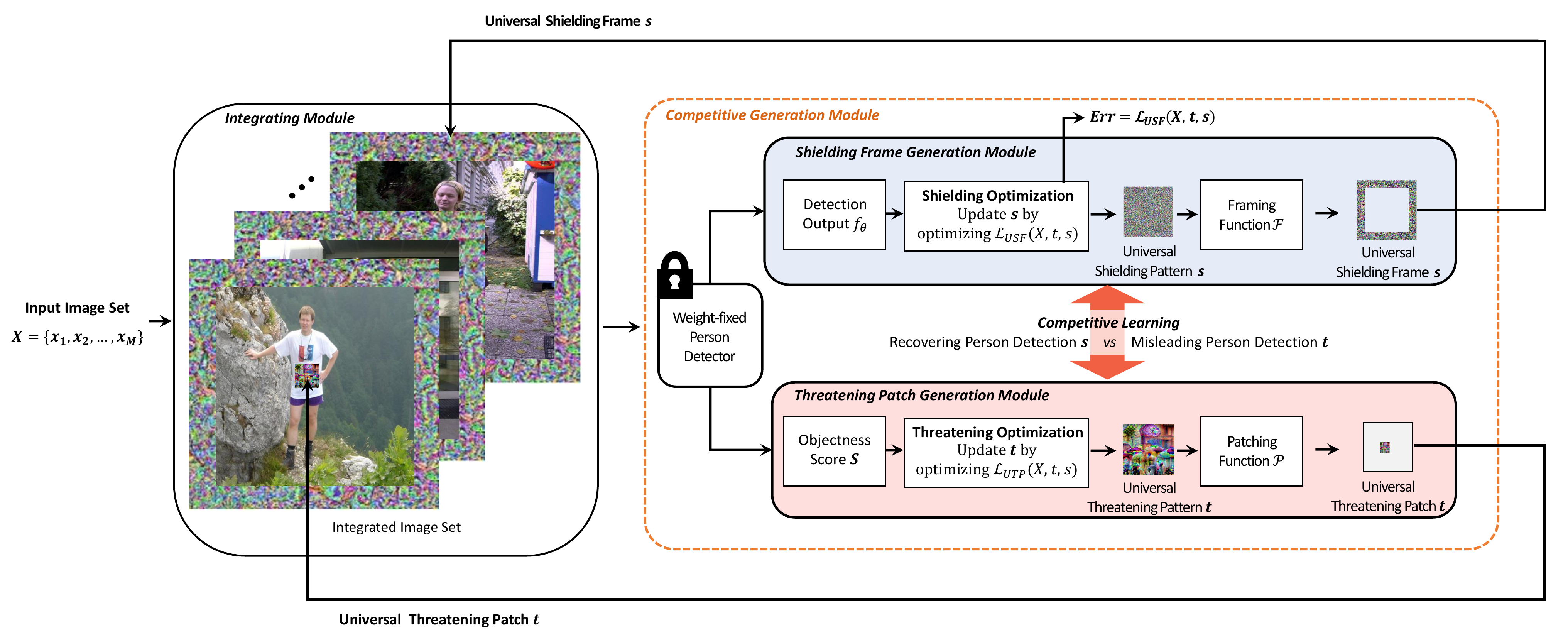}
	\caption{Detailed architecture of the competitive generation module. The architecture can include any weight-fixed detector (e.g., YOLO and Fast(er) R-CNN). The shielding frame generation module (blue, top) receives the detection output $f_\theta$ as an input and outputs the loss function $\mathcal{L}_{USF}$ and the universal shielding frame $s$. The threatening patch generation module (red, bottom) takes an objectness score $\textbf{S}$ as input and returns the universal threatening patch $t$ as output. Competitive learning between these two modules encourages better output $s,t$. $f_\theta$ is the detection output which is the combined vector including the classification score, the objectness score, and the bounding box location. $\mathcal{L}_{USF}$ and $\mathcal{L}_{UTP}$ are the loss functions that optimize USF and UTP, respectively.} 
	\label{fig:3}
\end{figure*}

\subsection{Competitive Game Between Adversarial Patch Attack and Defensive Frame in Person Detection} \label{sec:3.2}
We achieve the defensive effect of the proposed defensive frame by exploiting the benefit of the competitive learning process. The research field of adversarial attack and defense has grown rapidly due to its competitive nature. The arms race between adversarial attack methods and defense methods induces a significant breakthrough for each side. We use this competitive philosophy of adversarial attack and defense to generate a robust defensive frame. To do this, we create both an adversarial patch and a defensive frame at the same time in the defensive frame generation process. The adversarial patch and the defensive frame compete with each other. While the adversarial patch misleads the original prediction of person detection, the defensive frame recovers the original prediction. Each generation process is strongly encouraged by the competition until the generated defensive frame converges and reaches a balance between them (zero-sum game). The final objective of our proposed defensive frame is to recover the detection result of the original benign image $x$ given an input adversarial patched image $\PA(x,p)$ in the testing phase. As shown in Figure \ref{fig:2}, to avoid any misunderstanding with the adversarial patch $p$ in the inference stage, we name the adversarial patch in the generation stage as a "threatening patch". Similarly, we name the defensive frame against the threatening patch in the generation stage as a "shielding frame". Therefore, optimization of the shielding frame and optimization of the threatening patch compete with each other in the generation process.

\section{Proposed Universal Defensive Frame in Person Detection} \label{sec:3}
\subsection{Overview}
As we already mentioned in the previous section, we distinguished (shielding frame/threatening patch) in the generation stage from (defensive frame/adversarial patch) in the inference stage. Figure \ref{fig:2} shows that the overall generation process for a universal shielding frame (USF) $s$ consists of several modules. First of all, the input image set $X$ is integrated with the universal shielding frame $s_k$ and the universal threatening patch (UTP) $t_k$ in the integrating module. The shielding frame generation module generates and optimizes the universal shielding frame $s_X$ for the sub-image set $X$. The iteratively generated outputs of the shielding frame generation module are the error rate (Err) and the shielding frame $s_k$. In the threatening patch generation module, the universal threatening patch $t_k$ is also generated in the iteration. Next, the error condition controls the termination of iterative generation for optimized universal shielding frame according to a proper threshold of error rate $Err=\mathcal{L}_{USF}\left(x,t_{k},s_{k}\right)$. After optimizing for all sub-image sets in the entire dataset, we use the final optimized universal shielding frame $s^*$ as universal defensive frame $d$ to defend against adversarial patches $p$ in the testing phase.

Figure \ref{fig:3} describes the detailed architecture of the competitive generation module. The main objective is to create a more robust defensive frame by conducting competitive learning between two players: the shielding frame generation module and the threatening patch generation module. The see-saw game between these two competitive optimization steps boosts the adversarial robustness of the final universal shielding frame. In the competitive generation module, we can use any weight-fixed person detector (e.g., YOLO and Fast(er) R-CNN) to optimize the universal shielding frame. 



\subsection{Integrating Module for Person Detection} 
A defensive frame should be image-agnostic in real-world applications. Therefore, in the first integrating module, we designed a universal defensive frame (UDF) that is effectively defensive for every input image set. In other words, once generated, the same UDF can protect almost all sample images against adversarial patches.  Let $X=\{x_1,x_2,...,x_M\}$ be a subset of images sampled from the distribution $\mu$ where $\mu$ denotes the distribution of the training images. We iteratively update for sub-image set $X$ \cite{moosavi2017universal}. The main goal of our method is to optimize the defensive frame to reduce the difference between the adversarially patched input with the defensive frame and the original input image in the inference stage. In the generation stage, a sampled input image set $X$ is integrated with universal shielding frame $s$ and universal threatening patch $t$ in the image domain. The integrated image set is then used as the input for a weight-fixed person detector such as YOLO and Faster R-CNN. 

\subsection{Competitive Generation Module for Person Detection}
The competitive generation module incorporates two competing modules (Figure \ref{fig:3}). One is the shielding frame generation module, and the other is the threatening patch generation module. In this module, these two sub-modules compete with each other. The competition drives both of them to enhance their capability.

\subsubsection{Shielding Frame Generation Module for Detecting Person} 
Considering that adversarial patches \cite{thys2019fooling} are designed to mislead the original prediction of the person detector, we aim to add a defensive frame around all the inputs so that we can reduce the effect of adversarial patches in the inference stage.

In the generation stage, to defend against the threatening patch $t$ that fools person detection, we generate the shielding frame $s$. In this section, we define the formulation of the shielding frame and describe how to generate it. We consider a shielding frame $s$ that could alleviate the effect of a threatening patch $t$. Since the threatening patch shares the same objectives with the adversarial patch attack, the shielding frame $s$ is successful if the detection model maintains its original detection, even though there is a threatening patch $t$. Therefore, the objective problem can be defined as follows,

\begin{equation}
\label{eqn:3}
{\min\limits_{s}}\hspace{0.1cm} {\parallel{f_\theta}( \F (\PA(x,t),s)) - f_{\theta}(x)\parallel_{r}},
\end{equation}

\noindent where the framing function $\F(\cdot)$ transforms universal shielding patterns to a frame and places the universal shielding frame $s$ on the outside edge of the image. To optimize the universal shielding frame (USF), which recovers the original prediction regardless of images, we take the expectation with respect to a subset of images $X$. According to Equation (\ref{eqn:3}), we optimize the universal shielding frame $s$ by minimizing the following loss function ($\min\limits_{s} \mathcal{L}_{USF}$),

\begin{equation}
\label{eqn:4}
    \mathcal{L}_{USF}\left(X,t,s\right)={\E_{(X\sim\mu)}}\parallel {f_\theta}( \F (\PA(X,t),s))-f_{\theta}(X){\parallel}_{r}.
\end{equation}

\noindent where $X$ is the set of $M$ images sampled from the distribution $\mu$, $\PA(\cdot)$ is the patching function that places a threatening patch $t$ on the person in $X$ data points. Equation (\ref{eqn:4}) optimizes the universal shielding frame to minimize the difference between the prediction of ${f_\theta}( \F (\PA(X,t),s))$ and $f_{\theta}{(X)}$. Through the adversarial optimization process between the defensive frame and adversarial patch, this difference is gradually converged to local optima, where it asymptotically closes to zero to make it to be small population risk. Then, each proposed defensive frame can exist for every patch-based attack(adv-obj, adv-cloak, adv-T-shirt, naturalistic patch, etc).

\subsubsection{Threatening Patch Generation Module for Fooling Person Detection} 
In the threatening module, threatening patch $t$ is optimized to reduce the objectness score for fooling person detection. Thus, the threatening patch can be optimized by minimizing the following loss function (${\min\limits_{t} \mathcal{L}_{UTP}}$),

\begin{equation}
\label{eqn:5}
\mathcal{L}_{UTP}(X,t,s)={\E_{(X\sim\mu)}}[\max({\textbf{S}( \F (\PA(X,t),s)), 0)]}.
\end{equation}

\noindent Here, $X$, the patching function $\PA(\cdot)$, and the framing function $\F(\cdot)$ are the same as in Equation (\ref{eqn:4}). $\textbf{S}(\cdot)$ denotes the output vector of objectness scores for $X$ data points. By optimizing the above loss function, a universal threatening patch can be generated to fool person detection regardless of input images. 

\begin{figure*}[t]
	\centering
	    \includegraphics[width=0.75\linewidth]{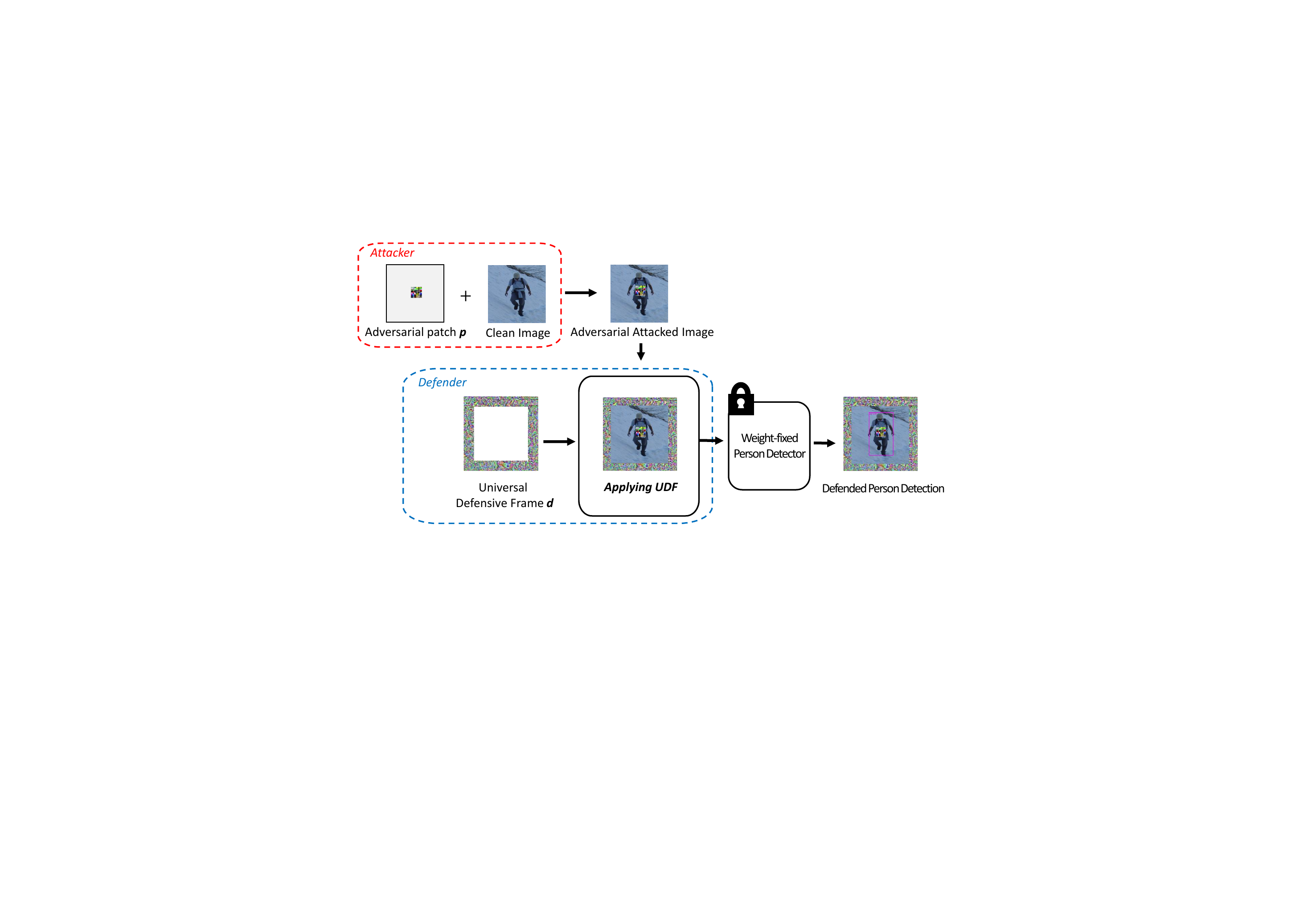}
	\caption{Person-detection defense process in the inference stage. The adversary can attack a clean image by adding an adversarial patch so that a physical adversarial patch fools the person detection. Our defense method can detect the person by applying UDF without re-training the person detector.} 
	\label{fig:4}
	\vspace{-0.5cm}
\end{figure*}

\subsubsection{Competitive Learning}
As shown in Figure \ref{fig:3}, the shielding module generates and updates the universal shielding frame $s$. On the other hand, the threatening module generates and updates the threatening patch $t$. The universal threatening patch tries to reduce the objectness score and misleads the original person detection. The universal shielding frame aims to recover the original person detection and reduces the adversarial effect of the universal threatening patch $t$. The threatening patch attempts to reduce the objectness score and misleads the original prediction in USF applied images. In contrast, $s$ has the constraint to recover the original prediction. With the changed $t$ decreasing the objectness score, the only way to recover the original prediction is to increase the objectness score of the person class for $s$ in equation (4). 
Through this cat-and-mouse game, the universal shielding frame can learn how to recover the decreased objectness score by threatening patch. This learned trait of enhancing the objectness score for person detection allows the universal shield frame to maintain the original prediction. We iteratively update the universal shielding frame and universal threatening patch according to the following updated equation,

\begin{equation}
\label{eqn:7}
    t_{k+1}=t_{k}-\mathrm{\nabla}_{t}  \mathcal{L}_{UTP} (X,t,s),
\end{equation}

\begin{equation}
\label{eqn:8}
    s_{k+1}=s_{k}-\mathrm{\nabla}_{s} \mathcal{L}_{USF} (X,t,s),
\end{equation}

\noindent where $k$ denotes the iteration step, the initial universal shielding frame and universal threatening patch ($s_{0}, t_{0}$) are set as Gaussian noise. We optimize Equation (\ref{eqn:7}) and (\ref{eqn:8}) step-by-step with the Adam \cite{kingma2014adam} algorithm. During the optimization, we fixed the weight parameter $\theta$ and only changed the values in the shielding frame and the threatening patch. After we optimized USF, we clipped it to assure that it is valid for images with real pixel values [0, 255]. The final optimized shielding frame $s^*$ becomes the universal defensive frame $d$.

\begin{algorithm}[!t]
\SetAlgoLined

\KwIn{Data point $X\sim \mu$, desired defense error rate $\delta$}
\KwOut{Universal defensive frame $d$ $\leftarrow$ $s^*$}
Initialize $s_0,t_0$ $\leftarrow$ \textit{Gaussian Noise}

\Repeat{training epoch $E$}{ 

\For{each data point $x\in{X}$}{
\For{$k=1,2,\dots,K_t$}{
$\Delta t=\nabla_{t} \mathcal{L}_{UTP}(x,t,s)$

$t_{k+1}=t_{k}-\Delta t$
}
}
\While{$Err(X) \geq \delta$}{
\For{\text{each data point $x\in{X}$}}{
\For{$k=1,2,\dots,K_s$}{
$\Delta s=\nabla_{s} \mathcal{L}_{USF}(x,t,s)$

${s}_{k+1}={s}_{k}-\Delta{s}$
}
}
}

}
\caption{Generating a universal defensive frame in person detection}
\label{alg:1}
\end{algorithm}

The detailed algorithm for generating a universal defensive frame is provided in Algorithm \ref{alg:1}. We use a greedy method to find the optimal $s^*$. The algorithm runs iteratively over the data points of $X$. At each iteration, we compute the $\Delta s$ to maintain the person detection of the current point. The greedy search terminates when $s$ is sufficient to keep the original person detection. We then repeat the optimization process until $Err(X)$ is smaller than the threshold $\delta$.

\begin{equation}
\label{eqn:9}
    Err(X) := \mathcal{L}_{USF}(X,t,s) < \delta.
\end{equation}

\noindent Since the universal shielding frame $s$ is optimized by considering the data distribution $\mu$, then it could be effective on most image samples. As such, the universal defensive frame $d$ could also be image-agnostic in the inference stage. The convergence analysis on competitive learning algorithm is in Appendix A.

\subsection{Defense in the Inference Stage} 
One of the benefits of our defensive frame in person detection is a relatively negligible cost in the testing phase. Figure \ref{fig:4} shows that the adversary can generate the adversarial patch $p$ on the weight-fixed person detector. After that, the adversary attaches the adversarial patch on a clean image and try to fool the person detector as if they had received a normal input image. In our defense method, the defenders do not need to re-train the person detector. Furthermore, defenders do not have to access to the original samples in the inference stage. In other words, once we optimize the UDF from the training dataset, there is no need to further optimize during the inference. Thus, there is no extra cost to use our defense method in the testing phase. Defenders just put the defensive frame $d$ on the outside edge of the image. The integrated input is used as an input for any weight-fixed person detector. Therefore, our method is sufficiently practical to use.

\section{Experiment}
\subsection{Experiment Setup}
\subsubsection{Person Detectors} 
We used two popular detectors to evaluate the proposed method: YOLO-v2 \cite{redmon2017yolo9000} and Faster R-CNN \cite{ren2015faster}. These person detectors are both pre-trained on the COCO dataset \cite{lin2014microsoft} that includes a person. The minimum detection threshold is set as 0.5 for both Faster R-CNN and YOLO-v2 by default. Since the YOLO network takes fixed image size, we can resize the image to $416\times416$ after applying UDF. In the Faster R-CNN, we pad the inputs so that the shape of the image becomes square. We use the public source code to generate an adversarial patch \footnote[1]{https://github.com/wangzh0ng/adversarial$\_$yolo2}. For the implementation, we adopt Pytorch 1.2 and CUDA 9.2 with a single GEFORCE GTX 1080Ti GPU. 

\begin{table*}[!t]
\centering
\caption{Defense performance on YOLO-v2 and Faster R-CNN with various adversarial patch attacks. The adversarial patches significantly decrease the AP score. However, by applying UDF, defense performance can be significantly improved.}
\begin{tabular}{cccccc}
\specialrule{.15em}{.1em}{.1em} 
\multirow{2}{*}{Dataset}                                                     & \multirow{2}{*}{Attack Method} & \multicolumn{2}{c}{YOLO-v2} & \multicolumn{2}{c}{Faster R-CNN} \\ \cline{3-6} 
                                                                             &                                & No Defense      & Ours       & No Defense         & Ours        \\ \hline
\multirow{4}{*}{INRIA Dataset}                                               & No Attack                      & 89.63           & 88.92      &           89.32         &    89.00         \\
                                                                             & Adv-Patch \cite{thys2019fooling}            & 18.80           & 51.96      &      49.21              &   69.32          \\
                                                                             & Adv-Cloak \cite{wu2020making}            &     14.81            &  43.55          &        38.50            & 65.52            \\
                                                                             & Adv-T-Shirt \cite{xu2020adversarial}          & 24.32                &       47.22     &          40.27          &       63.30      \\ \hline
\multirow{4}{*}{\begin{tabular}[c]{@{}c@{}}Collected\\ Dataset\end{tabular}} & No Attack                      & 97.58           & 96.93      &      96.81              &    96.57         \\
                                                                             & Adv-Patch \cite{thys2019fooling}            & 1.90            & 51.05      &           28.90         &    61.33         \\
                                                                             & Adv-Cloak \cite{wu2020making}            &   1.40              &        48.52    &               34.24     & 63.75            \\
                                                                             & Adv-T-Shirt \cite{xu2020adversarial}          &     19.53            &  49.93          &            39.00        &       61.40      \\ \specialrule{.15em}{.1em}{.1em} 
\end{tabular}
\label{tab:1}
\vspace{-0.2cm}
\end{table*}

\begin{table*}[!t]
\centering
\caption{Defense performance on YOLO-v2 and Faster R-CNN under adaptive attack setting. The adversarial patches are generated from UDF applied images.}
\begin{adjustbox}{max width=0.99\textwidth}
\begin{tabular}{cccccc}
\specialrule{.15em}{.1em}{.1em} 
\multirow{2}{*}{Dataset}                                                     & \multirow{2}{*}{Attack Method} & \multicolumn{2}{c}{YOLO-v2} & \multicolumn{2}{c}{Faster R-CNN} \\ \cline{3-6} 
                                                                             &                                & No Defense      &  Ours       & No Defense         &  Ours        \\ \hline
\multirow{3}{*}{INRIA Dataset}                                               
                                                                             & Adv-Patch \cite{thys2019fooling}             & 18.80           & 44.49      &      49.21              &   59.41          \\
                                                                             & Adv-Cloak \cite{wu2020making}             &     14.81            &  41.22          &        38.50            & 55.23            \\
                                                                             & Adv-T-Shirt \cite{xu2020adversarial}           & 24.32                &       44.71     &          40.27          &       54.82      \\ \hline
\multirow{3}{*}{\begin{tabular}[c]{@{}c@{}}Collected \\ Dataset \end{tabular}} 
                                                                             & Adv-Patch \cite{thys2019fooling}            & 1.90            & 45.32      &           28.90         &    48.92         \\
                                                                             & Adv-Cloak \cite{wu2020making}             &   1.40              &        40.18    &               34.24     & 52.68            \\
                                                                             & Adv-T-Shirt \cite{xu2020adversarial}           &     19.53            &  40.04          &            39.00        &       50.49      \\ \specialrule{.15em}{.1em}{.1em} 
\end{tabular}
\end{adjustbox}
\label{tab:4}
\vspace{-0.3cm}
\end{table*}

\subsubsection{Datasets}
To generate an adversarial patch and universal defensive frame, we used two datasets for person detection. 
\noindent \textbf{INRIA Dataset:} The first is the INRIA person dataset \cite{dalal2005histograms}. The INRIA person dataset is a publicly available dataset, and is for evaluating the fooling effect of the adversarial patch on a person detection algorithm \cite{thys2019fooling, wu2020making, xu2020adversarial}. It consists of 614 training images and 288 test images with various scenes. 

\noindent \textbf{Collected Dataset:} We collected our own person detection dataset which consists of a security entrance scene. The dataset is collected from two different outdoor scenes and one indoor scene with FLIR Pro R camera. The dataset contains videos on three different scenes: two outdoor scenes and one indoor scene. Each video takes one hour and was captured by a moving person. The resolution of the video is $512\times640$ ($Height \times Width$). After taking videos, we sampled images that capture persons. As a result, we obtained 130 images in the indoor scene, 500 images in the outdoor scene 1, and 187 images in the outdoor scene 2. From these images, we generated a training set by combining each indoor scene and outdoor scene 1. We then used the rest of outdoor scene 2 as a validation set. Therefore, we generated 630 training set and 187 validation set. We manually annotated all images after collecting the dataset.
\subsubsection{Adversarial Patch Setting} For the adversarial patch attack that fools the person detection, we used three different algorithms: Adv-patch~\cite{thys2019fooling}, Adv-T-shirt~\cite{xu2020adversarial}, and Adv-cloak~\cite{wu2020making}. The patch transformation function is utilized during the generation including resizing, scaling, random noise, and brightness changes. Unless otherwise noted, we used the same optimization parameter settings as each original paper. To optimize the adversarial patch, Adam optimizer was used with an initial learning rate of 0.03. After the empirical evaluations, we set the parameters $E$, $K_t$, and $K_s$ in Algorithm 1 as 15, 10, and 10, respectively.

\subsection{Quantitative Results of Universal Defensive Frame}

We verified that the universal defensive frame (UDF) can defend against adversarial patch attacks. To this end, we generated the three aforementioned adversarial patches that attacked the person detection \cite{thys2019fooling,wu2020making,xu2020adversarial}. We then applied the generated UDF to the image and computed the average precision (AP) score to verify the defense performance. The AP score balances the trade-off between precision and recall. To generate UDF, we set the frame width $w=80$. Table \ref{tab:1} shows the experimental results on two different datasets and two types of detection models. 
As shown in Table \ref{tab:1}, for the clean image (No Attack), the YOLO-v2 model shows an AP score of 89.63. When we applied adversarial patches, the AP score is reduced to 18.80 for the Adv-Patch attack, 14.81 for the Adv-Cloak attack, and 24.32 for the Adv-T-Shirt attack. The AP score is improved versus the no defense results when applying UDF to the image (Ours). Table \ref{tab:1} shows that the proposed approach improves the AP score by 33.16, 27.74, and 22.90 on Adv-Patch, Adv-Cloak, and Adv-T-Shirts attacks respectively. Furthermore, the AP score is only reduced by 0.71 in the case of the clean image (No Attack). The experimental results show that UDF can suppress the effect of the adversarial patch and defend against the adversarial patch attack while maintaining its original prediction.

Furthermore, we generated UDF on the Faster R-CNN model and obtained similar results. As shown in the table, it improves the AP score by 20.11, 27.02, and 23.03 on Adv-Patch, Adv-Cloak, and Adv-T-Shirt attacks respectively. It also keeps the original AP score on clean images. Similar results are presented on the collected dataset. The experimental results demonstrate that the proposed method can defend against an adversarial patch attack. Furthermore, the proposed method can be general and flexible enough to be applicable to both one-stage and two-stage detection networks.

\begin{figure*}[]
	\centering
	    \includegraphics[width=0.99\linewidth]{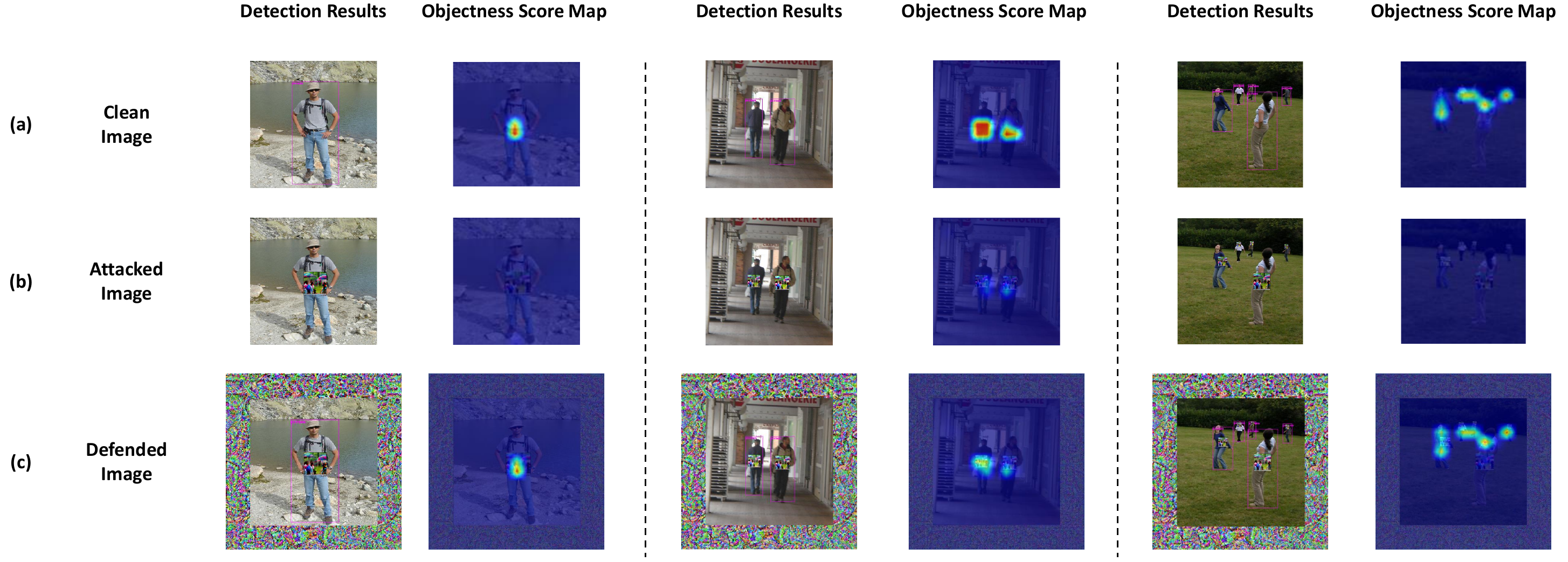}
	\caption{Comparison of objectness score map for YOLO-v2 network. (a) input image and objectness score map for clean image; (b) input image and objectness score map for attacked image with the adversarial patch; (c) input image and objectness score map for defended image with universal defensive frame.} 
	\label{fig:6}
\end{figure*}

\begin{table*}[t!]
\centering
\caption{Defense performance on recently proposed SOTA person detectors (ALFNet, CSP, LPRNet, and ACSP) with various adversarial patch attacks on two datassets.}
\resizebox{0.9\textwidth}{!}{%

\begin{tabular}{cccccccccc}

\specialrule{.15em}{.1em}{.1em}
\multirow{2}{*}{Dataset}           & \multirow{2}{*}{Attack Method} & \multicolumn{2}{c}{ALFNet\cite{liu2018learning}} & \multicolumn{2}{c}{CSP\cite{liu2019high}} & \multicolumn{2}{c}{LPRNet\cite{kim2021robust}} & \multicolumn{2}{c}{\mbox{ACSP\cite{wang2020detection}}} \\ \cline{3-10} 
                                   &                                & No Defense     & Ours      & No Defense    & Ours    & No Defense     & Ours     & No Defense     & Ours   \\ \hline
\multirow{4}{*}{Inria Dataset}     & No Attack                      & 90.21          & 88.64     & 93.74         & 91.14   & 93.02          & 90.97    & 94.37      & 92.22 \\
                                   & Adv-Patch \cite{thys2019fooling}             & 24.58          & 53.33     & 35.01         & 52.42   & 33.21          & 49.07   &36.27  &54.17    \\
                                   & Adv-Cloak \cite{wu2020making}             & 19.57          & 42.40      & 27.72         & 47.22   & 27.12          & 44.75     &30.01  &49.31 \\
                                   & Adv-T-shirt \cite{xu2020adversarial}           & 20.31          & 45.79     & 31.37         & 50.55   & 30.51          & 43.29    &32.08   &52.73 \\ \hline
\multirow{4}{*}{Collected Dataset} & No Attack                      & 95.83          & 93.22     & 97.63         & 95.28   & 96.21          & 94.15     &97.98      &95.52 \\
                                   & Adv-Patch \cite{thys2019fooling}             & 15.11          & 47.11     & 32.21         & 50.76   & 28.20           & 49.59     &35.21   &53.74\\
                                   & Adv-Cloak \cite{wu2020making}             & 10.33          & 43.74     & 22.38         & 39.74   & 19.53          & 47.87     &26.42   &43.17\\
                                   & Adv-T-shirt \cite{xu2020adversarial}           & 5.42           & 45.93     & 27.19         & 45.50    & 25.86          & 41.49     &29.34     &48.89\\ \specialrule{.15em}{.1em}{.1em}
\end{tabular}%
} \label{tab:person}
\end{table*}

\subsection{Adaptive Patch Attack}
One may wonder if attackers are aware that a universal defensive frame will be added before the detection, whether they can design a better adversarial patch. This defense-aware adversarial patch is an adaptive patch attack in the white-box attack setting. Our universal defensive frame is sufficiently effective against this adaptive patch attack (see Appendix A). Following the referred paper~\cite{athalye2018obfuscated}, in order to verify whether the defense strategy is Gradient Masking (false robustness) or not, it is necessary to verify whether it can defend against adaptive attack. In other words, the attacker can accurately know the existence and value of the UDF, generate adversarial patches from the image already applied with the UDF, and then evaluate the defense performance.


To demonstrate the effectiveness of the proposed UDF under the adaptive attack setting, we conduct the experiment. Empirical evidence supports that the proposed universal defensive frame is effective against the adaptive patch attack. To generate the adaptive patch attack, we first optimized the UDF ($d$) with training dataset. After optimization, we used a fixed $d$ without further optimization and without changing the value of $d$. We then applied $d$ to the test image ($x_n$) where $n$ denotes the index of test image and the UDF applied image can be written as $\mathcal{F}(x_n,d)$. After applying the UDF to the test images, we generated an adversarial patch $({p_n}^{adaptive})$ for all $\mathcal{F}(x_n,d)$ and used the adversarial patched images for evaluation. In other words, ${p_n}^{adaptive}$ indicate the adaptive patch that attacks UDF applied image $\mathcal{F}(x_n,d)$. Note that, since the adversarial patch is generated on UDF applied images, it can be regarded that the attacker generates adversarial patches by knowing the existence of the UDF and using the value of UDF accurately. The results with an adaptive adversarial patch are shown in Table \mbox{\ref{tab:4}}. Table \mbox{\ref{tab:4}} shows that the proposed universal defensive frame is still effective under adaptive attacks on the YOLO-v2 and the Faster R-CNN detection.

\begin{table*}[t]
\centering
\caption{Defense performance comparison with recently proposed defense methods against adversarial patch attacks.}
\resizebox{0.9\textwidth}{!}{%
\begin{tabular}{cccccccc}
\specialrule{.15em}{.1em}{.1em}
\multirow{2}{*}{Dataset}           & \multirow{2}{*}{\begin{tabular}[c]{@{}c@{}}Defense\\ Method\end{tabular}} & \multicolumn{3}{c}{YOLO-v2}                      & \multicolumn{3}{c}{Faster-RCNN}                  \\ \cline{3-8} 
                                   &                                                                           & Adv-Patch      & Adv-Cloak      & Adv-T-shirt    & Adv-Patch      & Adv-Cloak      & Adv-T-shirt    \\ \hline
\multirow{3}{*}{Inria Dataset}     & LGS\cite{naseer2019local}                                                                       & 30.32          & 25.11          & 29.52          & 52.11          & 43.02          & 41.65          \\
                                   & FNC\cite{yu2021defending}                                                                       & 38.59          & 29.66          & 29.8           & 55.31          & 48.27          & 44.01          \\
                                   & Ours                                                                      & \textbf{44.71} & \textbf{41.22} & \textbf{44.71} & \textbf{59.41} & \textbf{55.23} & \textbf{54.82} \\ \hline
\multirow{3}{*}{Collected Dataset} & LGS\cite{naseer2019local}                                                                       & 26.4           & 24.52          & 29.34          & 32.53          & 39.75          & 40.26          \\
                                   & FNC\cite{yu2021defending}                                                                       & 36.22          & 29.73          & 30.72          & 35.20           & 42.89          & 43.99          \\
                                   & Ours                                                                      & \textbf{45.32} & \textbf{40.18} & \textbf{40.04} & \textbf{48.92} & \textbf{52.68} & \textbf{50.49} \\ \specialrule{.15em}{.1em}{.1em}
\end{tabular}%
} \label{tab:defense}
\end{table*}

\begin{table}
\centering
\caption{Comparison with recent defense against adversarial patch attack in person detection for the INRIA dataset.}\label{tab:3}
\begin{tabular}{ccc}
\specialrule{.15em}{.1em}{.1em}
\multirow{2}{*}{Attack Method} & \multicolumn{2}{c}{Defense Method}                  \\ \cline{2-3} 
                                        & \multicolumn{1}{c}{Grad-Defense\cite{saha2020role}} & Ours           \\ \hline
No Attack                               & \multicolumn{1}{c}{\textbf{89.07}}         & {88.92}          \\ 
Adv-Patch \cite{thys2019fooling}                     & \multicolumn{1}{c}{37.27}                  & \textbf{51.96} \\ 
Adv-Cloak \cite{wu2020making}                     & \multicolumn{1}{c}{43.32}                  & \textbf{43.55} \\ 
Adv-Tshirt \cite{xu2020adversarial}                    & \multicolumn{1}{c}{41.72}                  & \textbf{47.22} \\ 
\specialrule{.15em}{.1em}{.1em}
\end{tabular}
\end{table} 

\subsection{Objectness Score Map Analysis}
According to previous papers \cite{thys2019fooling,xu2020adversarial, wu2020making}, the adversarial patches reduce the objectness score. To defend against the adversarial patch, the universal defensive frame suppresses the effect of the adversarial patch and then maintains the clean prediction. Figure \ref{fig:6} shows the objectness score maps of YOLO v2 according to different input types. Figure \ref{fig:6} (a) shows the objectness score maps with a clean image. As shown in the figure, the objectness score map is highly activated around the person, and thus it detects the person correctly. Figure \ref{fig:6} (b) shows the objectness score maps with the adversarially patched image. In contrast to the results of the clean image, when applying the adversarial patch, the activation of the objectness score map is reduced. Therefore, the detector fails to detect the person. Figure \ref{fig:6} (c) shows the objectness score maps when applying the universal defensive frame on the adversarially patched image. The model successfully detected the person even though the image is attacked by the adversarial patch. Also, as shown in the objectness score map, the original prediction is recovered similar to the score map of clean image. Moreover, even under the circumstance that adversarial patches are attached to multiple people of different locations and sizes (second and third columns of Figure \ref{fig:6}), UDF can recover the objectness score map and make the network correctly detect persons. Therefore, it can be interpreted that the universal defensive frame could suppress the effect of the adversarial patch and then recover its clean prediction.

\subsection{Defense Performance on Person Detection Specified Model}
In this section, we prove the effectiveness of the proposed universal defensive frame on the person detection specified detection models on two datasets. To this end, we chose the recently proposed state-of-the-art person detectors (ALFNet\mbox{\cite{liu2018learning}}, CSP\mbox{\cite{liu2019high}}, LPRNet\mbox{\cite{kim2021robust}}, and ACSP\mbox{\cite{wang2020detection}}) to compare the result. Table \mbox{\ref{tab:person}} shows the defense performance on person detection models. As shown in the table, SOTA person detectors are vulnerable to adversarial patch attacks. In Table~\mbox{\ref{tab:person}}, for the clean image (No Attack), ALFNet, CSP, LPRNet, and ACSP show AP score of 90.21, 93.74, 93.02, and 94.37, respectively. When, we applied adversarial patch (Adv-Patch\mbox{\cite{thys2019fooling}}), the AP score reduced to 24.58, 35.01, 33.21, and 36.27 respectively. However, when we apply the UDF, the AP score is increased to 53.33, 52.42, 49.07, and 54.17 respectively. Similar results could be verified for other patches as well. Therefore, the results can be interpreted and prove that the proposed UDF can effectively defend against adversarial patches on person detectors.

\subsection{Comparison with Other Defense Methods}
In this section, we verify the effectiveness of the proposed method by comparing it with adversarial patch defense method for detector. For the comparison, we reproduced the public official code of \cite{saha2020role} \footnote[2]{https://github.com/UMBCvision/Contextual-Adversarial-Patches}. \cite{saha2020role} exploited spatial-context information to reduce the effect of adversarial patch on YOLO-v2 detector. Table \ref{tab:3} shows the experiment results according to various adversarial patches. As shown in the table, we verify that the proposed method is more robust than \cite{saha2020role}. For the computational efficiency, our proposed method only took five hours to optimize the universal defensive frame with a single GPU, GEFORCE GTX 1080Ti. Whereas \cite{saha2020role} took three days to train the robust enough model with the same performance in Table \ref{tab:3}. Note that the proposed method does not need to re-train the model, which causes a marginal computational overhead. In other words, the universal defensive frame is flexible enough to be applicable to existing weight-fixed person detection models without changing the parameters. Therefore, our defense method outperforms the existing defense method in terms of computational efficiency as well.

Furthermore, we compared recently proposed adversarial patch defense methods applicable to various tasks. We compared the Local Gradient Smoothing (LGS)~\cite{naseer2019local} and the Feature Norm Clipping (FNC)~\cite{yu2021defending} against adaptive patch attacks. Table \ref{tab:defense} demonstrates that the proposed universal defensive frame shows superior performance than SOTA adversarial patch defense methods on both YOLO-v2 and Faster R-CNN network. The proposed UDF improved the defense performance by 31.56\% compared to FNC, and 55.06\% compared to LGS with the YOLO-v2 framework on average. Similarly, the proposed UDF increased the defense performance by 19.24\% compared to FNC, and 28.97\% compared to LGS with the Faster-RCNN framework on average. Similar tendency of defense performance was maintained between two different datasets (Inria and Collected dataset).

\begin{figure*}
	\centering
	    \includegraphics[width=0.65\textwidth]{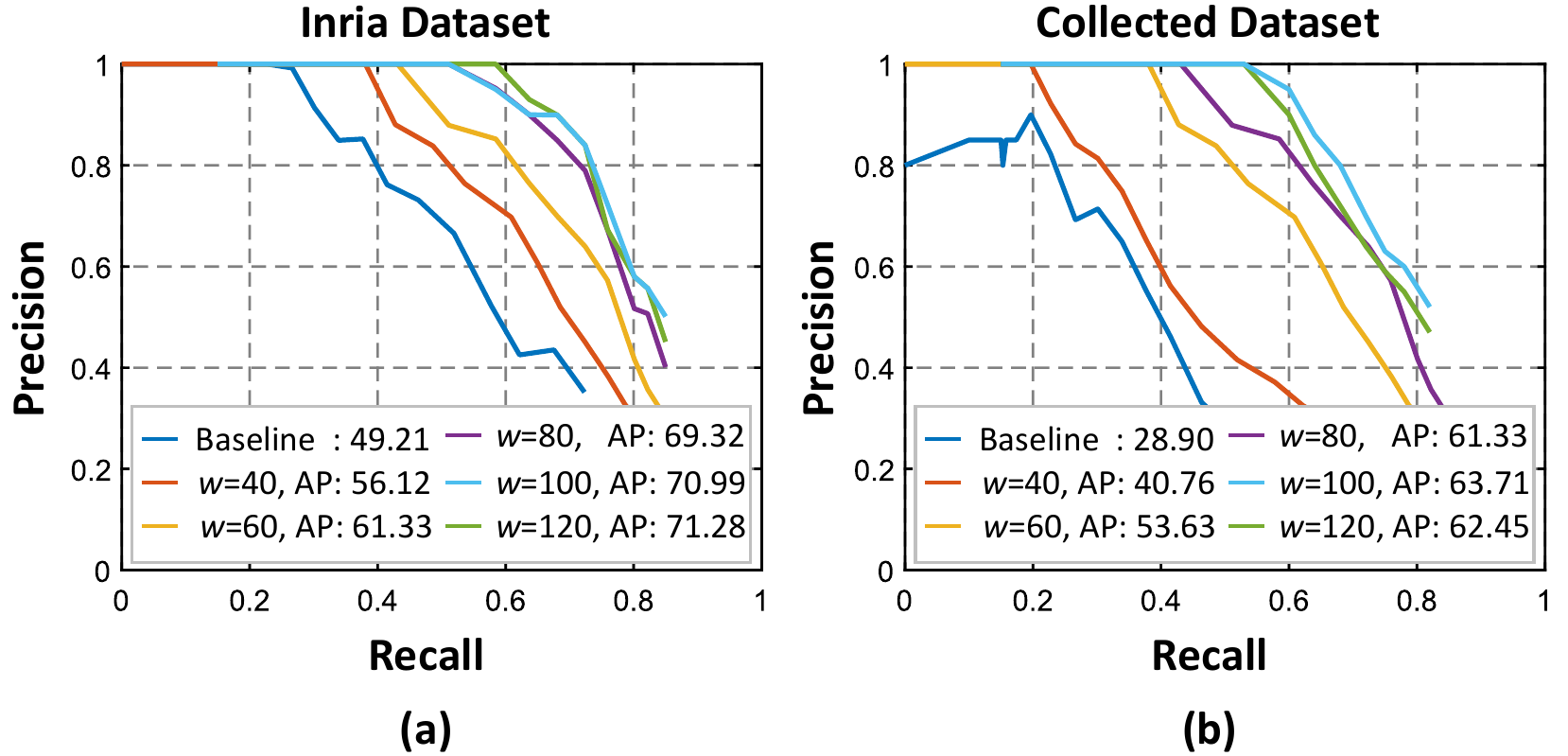}
	\caption{Precision-recall curves according to the frame width of UDF (\textit{w}=40, 60, 80, 100, 120) when the adversarial patch~\cite{thys2019fooling} is applied. The defense performance increases as the frame width w increases. Then, the defense performance is almost saturated after \textit{w}=80. (a) denotes the results on INRIA dataset and (b) denotes the results on Collected dataset.}
	\label{fig:5}
	\vspace{-0.2cm}
\end{figure*}

\begin{figure}[t]
	\centering
	    \includegraphics[width=0.99\linewidth]{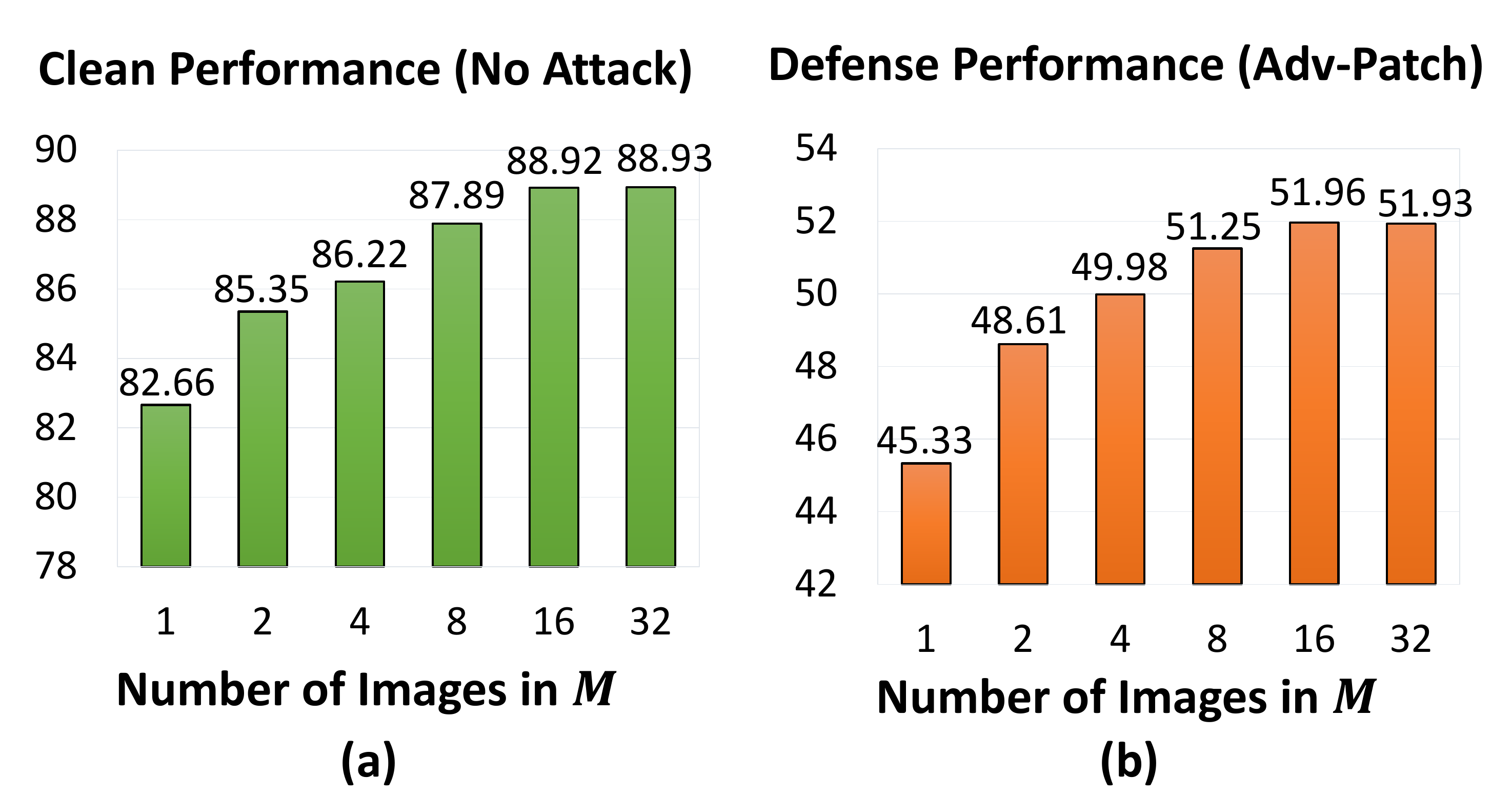}
	\caption{Detection performance according to the number of images $M$ on INRIA dataset. (a) clean image performance; (b) defense performance against adversarial patch.}
	\label{fig:7}
	\vspace{-0.5cm}
\end{figure}

\subsection{Detailed Analysis on Frame Width}
In this section, we analyze the effect of frame width \textit{w}. For the analysis, we use the adversarial patch from\cite{thys2019fooling}, and then change the frame width of UDF from 40 pixels to 120 pixels at 20 pixel intervals (\textit{w}=40, 60, 80, 100, 120) on Faster R-CNN. Figure \mbox{\ref{fig:5}} describes Precision-Recall curves on the INRIA dataset (a) and the collected dataset (b). The AP score increases as the frame width \textit{w} increases. In other words, we can increase the defense level by extending the frame width \textit{w}. Then, the defense performance is saturated after frame width 80. In other words, as the frame size increases, the performance saturates and does not degrade. Considering the trade-off between computation burden and the performance increase, we set the frame width \textit{w}=80 on other experiments.

Also, just as defense performance increases as the size of UDF increases, adversarial patches can induce the defender to have a larger frame width by increasing the size of the patches. However, since the size of the adversarial patch is limited to the size of the person, it is unnatural to increase the size of a patch indefinitely. If the size of the patch is increased by 1.5 times, the patch will cover more than the upper body of the person and exceed the bounding box. Therefore, the size of the adversarial patch can be increased up to about 1.5 times. To verify the effectiveness of the proposed method using large-sized adversarial patches, we perform an experiment by increasing the patches by 1.5 times. In our experiment, we achieved 44.72\% defense performance with UDF ($\textit{w}=80$). Compared with no defense (10.82\%), our proposed method is still effective and increases the defense performance by 33.9\% under the increased patch size situation.

\subsection{Ablation Study}
\subsubsection{Number of images M in the sub-image set X}
Figure \ref{fig:7} shows the defense performance of UDF against adversarial patch attack~\cite{thys2019fooling} with different number of images $M$ in the sub-image set $X$. As shown in the figure, when the size of subset image is small, the increase in defense performance is marginal. As the size of $M$ increase, the defense performance is increased. Then, the defense performance is saturated at $M=16$. Therefore, when we set $M=16$, it sufficiently defends against the adversarial patch attack on both datasets.

\subsubsection{Number of iterations $K_t$ and $K_s$}
Table \ref{tab:5} shows the defense performance of UDF against adversarial patch attack~\cite{thys2019fooling} with different number of iterations ($K_s$ and $K_t$). In this case, we fixed epochs $E$ as 15 in the experiment. Considering the concept of competitive learning, we also set the number of iterations of attack $K_t$ and defense $K_s$ as the same number. The defense performance increased when $K_t$ and $K_s$ increased. It was then saturated at 10 iterations. Therefore, we optimized UDF with $K_t=K_s=10$.

\subsubsection{Number of epochs $E$}
Table \ref{tab:6} demonstrates the defense performance of UDF against adversarial patch attack~\cite{thys2019fooling} with different number of epochs $E$. The defense performance of UDF increased with the number of epochs $E$. Then, the defense performance converged at $E=15$. Therefore, we set the training epoch $E$ as 15 for optimal choice.

\subsubsection{Error threshold $\delta$}
Table \ref{tab:8} shows the defense performance change with the error threshold $\delta$. The detection accuracy is diminished as the threshold is loosened (threshold increased). This is because the weakening of the threshold means that the universal defensive frame is optimized with a large error rate. Considering this empirical evaluation, we set the error threshold as 0.4.

\begin{table}[!t]
\caption{Defense performance change according to the number of iterations ($K_s$ and $K_t$).}
\begin{adjustbox}{max width=0.48\textwidth}
\begin{tabular}{cccccccc}
\specialrule{.15em}{.1em}{.1em} 
Iterations & 1   & 2     & 5    & 8    & 10    & 12    & 15    \\ \hline
AP score   & 40.94 & 43.81 & 47.79 & 49.67 & 51.96 & 51.93 & 51.95 \\ \specialrule{.15em}{.1em}{.1em} 
\end{tabular}
\end{adjustbox}
\label{tab:5}

\end{table}
\begin{table}[!t]
\caption{Defense performance change according to the number of epochs ($E$).}
\begin{adjustbox}{max width=0.48\textwidth}
\begin{tabular}{cccccccc}
\specialrule{.15em}{.1em}{.1em} 
Epochs   & 1     & 3    & 5    & 10   & 15   & 25   & 50  \\ \hline
AP score & 36.41 & 44.32 & 48.96 & 50.47 & 51.96 & 52.03 & 51.99 \\ \specialrule{.15em}{.1em}{.1em} 
\end{tabular}
\end{adjustbox}
\label{tab:6}

\end{table}

\begin{table}[!t]
\caption{Defense performance change according to error threshold ($\delta$).}
\begin{adjustbox}{max width=0.48\textwidth}
\begin{tabular}{ccccccccc}
\specialrule{.15em}{.1em}{.1em} 
Threshold    & 0.2   & 0.3   & 0.4   & 0.5   & 0.6   & 0.7   & 0.8  & 0.9   \\ \hline
AP score & 51.94 & 51.97 & 51.96 & 48.73 & 45.63 & 38.49 & 31.6 & 21.24 \\ \specialrule{.15em}{.1em}{.1em} 
\end{tabular}
\end{adjustbox}
\label{tab:8}

\end{table}

\begin{table}[t]
\caption{Run-time comparison (ms) of the proposed defense with YOLO-v2 and Faster R-CNN detectors.}
\begin{adjustbox}{max width=0.48\textwidth}
\begin{tabular}{ccccc}
\specialrule{.15em}{.1em}{.1em} 
              & \multicolumn{2}{c}{YOLO-v2} & \multicolumn{2}{c}{Faster R-CNN} \\ \cline{2-5} 
              & No Defense      & Ours      & No Defense         & Ours        \\ \hline
Run-time(ms/img) & 23.1            & 28.3      & 70.0               & 83.1        \\ \specialrule{.15em}{.1em}{.1em} 
\end{tabular}
\end{adjustbox}
\label{tab:7}
\end{table}

\subsection{Run-time Analysis}
The proposed universal defensive frame is pre-processing defense in the input domain. Thus, the inference time can increase. In this section, we compare the inference time with different detection network. To compare the run-time, we implement the prediction with a single GEFORCE GTX 1080Ti. Table \ref{tab:7} demonstrates the run-time comparison between the YOLO-v2 and Faster R-CNN network. At 80 pixels, the inference time is slowed down by about 13.1 and 5.2 milliseconds for Faster R-CNN models and YOLO models respectively. As shown in the table, since our method simply attaches the universal defensive frame in the input domain, the run-time does not increase so much. Therefore, the proposed method is time-efficient in practical application and sufficient to implement it in real-time.


\subsection{Discussion}
\subsubsection{Training the model with adversarial patched person data}
One of the ways to improve the detector's defense performance is training the model with adversarial patched (Adv-Patch\cite{thys2019fooling}) person data. On the other hand, the purpose of the proposed method is to increase the defense performance of the given detector by adding UDF to input image. Therefore, our proposed method can further improve defense performance by applying UDF to trained models including adversarial patches to training data. To verify this, we conduct an experiment to prove the effectiveness of UDF on the trained model including adversarial patches to training data. To this end, we train the detector with two datasets (Inria and Collected) including adversarially patched person. Table \ref{tab:at} shows the experiment results on the trained model including adversarial patches to training data. As shown in the table, if the model is trained with adversarial patches it could defend against adversarial patch (without UDF) compared with the conventionally trained model. Then, we could further improve the defense performance by applying UDF to input image (with UDF).

\begin{table}[t]
\caption{Defense performance of the model (Faster R-CNN) trained by including the adversarial patch in the training data against adaptive patch attack.}
\resizebox{0.98\linewidth}{!}{%
\begin{tabular}{cccc}
\specialrule{.15em}{.1em}{.1em}
\multirow{2}{*}{Dataset}           & \multirow{2}{*}{Attack Method} & \multicolumn{2}{c}{Trained   with Adversarial Pattern} \\ \cline{3-4} 
                                   &                                & without UDF                & with   UDF                \\ \hline
\multirow{3}{*}{INRIA Dataset}     & Adv-Patch                      & 54.21                      & 63.51                     \\
                                   & Adv-Cloak                      & 53.11                      & 59.82                     \\
                                   & Adv-T-shirt                    & 49.75                      & 60.72                     \\ \hline
\multirow{3}{*}{Collected Dataset} & Adv-Patch                      & 45.32                      & 56.33                     \\
                                   & Adv-Cloak                      & 40.67                      & 57.12                     \\
                                   & Adv-T-shirt                    & 44.61                      & 54.28                     \\ \specialrule{.15em}{.1em}{.1em}
\end{tabular}%
} \label{tab:at}
\end{table}

\subsubsection{Future work}
The goal of the proposed method is to optimize a defensive pattern that can protect the person class from the adversarial pattern which makes the person be hidden. According to the formulation, the defensive pattern is not solely applicable to the task of person detection. Learnable defensive pattern can be generalized to general object detection. We plan to extend our work to other target objects in future work.

\section{Conclusion}
In this paper, we showed the usefulness of the proposed defense method against adversarial patch attack in person detection. We devised universal defensive frame of person detection, which alleviated the effect of adversarial patch attack. To generate the defensive frame in person detection, we used a novel competitive learning framework between two competitive parties in person detection: shielding and threatening person-detection. The optimized defensive frame was applied to images to defend against adversarial patch attack in person detection. The proposed universal defensive frame is effective in real-world applications because it does not require a re-training of the existing person detector with heavy computational cost. Experimental results demonstrated that the proposed approach effectively defended against adversarial patch attack. We believe that the proposed universal defensive frame is practically useful in the person-detection defense against adversarial patch attack.

\appendices
\section{Theoretical Analysis}
\subsection{Analysis on the convergence of competitive learning}

To verify the convergence of the competitive learning algorithm, we adopt a simplified denotation for linear calculation. For simplicity, we also assume that the attack is computed by a gradient ascent on the loss of network outputs, which is a different assumption in Equations (5) and (6) in the manuscript. We define the zero-valued framing function ${F_{0}}(x)$ which adds a zero-valued frame border outside the input $x$. The size of the frame induced by ${F_{0}}(x)$ is exactly the same with the universal shielding frame. From this, if the universal threatening patch $t$ is added to the input image $x$, we can denote the new input as $x'=F_{0}(x+t)$, and we can also denote the new input $x'$ with universal shielding frame $s$ as $x''=x'+s=F_{0}(x+t)+s$.

We can also define $\mathcal{L}(f(a),f(b)) = \parallel f(a) - f(b)\parallel$. Then, we can borrow the iterative feed-forward process of shielding frame $s_k$ in Algorithm 1 as $\textbf{s}(x''_{k}) = \mathcal{L}(f(x''_{k}), f(x)) = \mathcal{L}(f(x''_{k}), f(x))$ where $x''_{k}=x'+s_{k} = F_{0}(x+t_k)+s_{k}$. The function $f$ is the detection result of the detector and the function $\textbf{s}(\cdot)$ is the feed-forward update of universal shielding frame by reducing the loss of difference between the detection result of shielded input $f(x''_{k})$ and original detection $f(x)$. The corresponding first order derivative with respect to the input as $\partial\textbf{s}(x''_{k})/\partial{x''_{k}}$. 

From this condition of new input $x'$ with universal threatening patch $t$, then the unified detector function $U(x')$ of deep networks and the shielding frame as an input processing method $S$ can be formulated as $U = f\circ{S(x')}$.

\begin{equation}
    U = f\circ{S(x')}.
\end{equation}

\noindent where $S$ is our input processing method, which is the universal shielding frame with

\begin{equation}
    S(x')=x'+\sum_{k=1}^{K_s}\frac{\partial\textbf{s}(x''_{k})}{\partial{x''_{k}}}.
\end{equation}

\begin{equation}
    U(x')=f(x'+\sum_{k=1}^{K_s}\frac{\partial\textbf{s}(x''_{k})}{\partial{x''_{k}}}) = f(x''_{k}).
\end{equation}

\begin{table}[t!]
\caption{Empirical evaluations on the magnitude of the first term and the magnitude of the second term in Equation (12) according to the change of the defense iterations $K_s$ with the same framework in Algorithm 1. First term denotes $\frac{\partial{\mathcal{L}}(f(x''_{K_s}),f(x))}{\partial{x''_{K_s}}}$ and Second term denotes $\sum_{k=1}^{K_s}\frac{\partial^2\textbf{s}(x''_{k})}{\partial{x''_{k}}\partial{x'}}$.}
\begin{adjustbox}{max width=0.49\textwidth}
\begin{tabular}{cccccc}
\specialrule{.15em}{.1em}{.1em} 
$K_s$              & 1        & 5        & 10       & 50       & 100      \\ \hline
First term  & 0.344    & 0.340    & 0.312    & 0.328    & 0.323    \\
Second term & 8.235e-5 & 8.662e-5 & 8.951e-5 & 9.248e-5 & 9.480e-5 \\ \specialrule{.15em}{.1em}{.1em} 
\end{tabular}
\end{adjustbox}
\label{tab:appen}
\end{table}


We can simplify the problem by considering only one incoming input image $x$, not an image set $X$. And we omit the projection operation during the attack process which will not affect our final conclusion. Then, the threatening patch for image with shielding frame should compute the first order derivative of the above input processing aware function with respect to the new incoming input $x'$

\begin{equation}
\begin{aligned}
    \frac{\partial{\mathcal{L}}(U(x'),f(x))}{\partial{x'}} &= \frac{\partial{\mathcal{L}}(f(x''_{K_s}),f(x))}{\partial{x'}} \\
    &= \frac{\partial{\mathcal{L}}(f(x''_{K_s}),f(x))}{\partial{x''_{K_s}}} \cdot \frac{\partial{x''_{K_s}}}{\partial{x'}} \\
    &= \frac{\partial{\mathcal{L}}(f(x''_{K_s}),f(x))}{\partial{x''_{K_s}}} \cdot \frac{\partial{(x'+\sum_{k=1}^{K_s}\frac{\partial\textbf{s}(x''_{k})}{\partial{x''_{k}}})}}{\partial{x'}} \\
    &= \frac{\partial{\mathcal{L}}(f(x''_{K_s}),f(x))}{\partial{x''_{K_s}}} \cdot (1+ \sum_{k=1}^{K_s}\frac{\partial^2\textbf{s}(x''_{k})}{\partial{x''_{k}}\partial{x'}}).
\end{aligned}
\end{equation}

Empirical evidence demonstrates that the term $\sum_{k=1}^{K_s}\frac{\partial^2\textbf{s}(x''_{k})}{\partial{x''_{k}}\partial{x'}}$ is extremely marginal whereas its computation can be very expensive (See Table \ref{tab:appen}). Then, the variation of second derivative term according to the $K_s$ is also marginal for a sufficiently large $K_s$. This is because if $K_s$ is increased in a sufficiently large enough, then the corresponding second derivative term (the fluctuation of the gradient according to the infinitesimal change in the input) is also significantly decreased as the proposed UDF becomes robust~\cite{gupta2022improved}. In this case, the amount of accumulation becomes negligible in a large $K_s$. Therefore, we can omit this part and obtain:

\begin{equation}
\label{eq:14}
    \frac{\partial{\mathcal{L}}(U(x'),f(x))}{\partial{x'}} \approx \frac{\partial{\mathcal{L}}(f(x''_{K_s}),f(x))}{\partial{x''_{K_s}}}.
\end{equation}

This means, on each attacking step of the universal threatening patch for the universal shielding frame, attacking process should first consider the whole iteration in Algorithm 1 and then apply the computed gradient with respect to $x''_{K_{s}}$, instead of computing the derivative with respect to $x'$. In other words, the gradient update for the upcoming input ($x'$) for the threatening patch is approximately the same with the gradient update for the input ($x''_{K_{s}}$) in the  final iteration $K_s$. This shows the possible existence of the balance and the convergence for the competitive learning algorithm.

\subsection{Analysis on the adaptive attack}

In the evaluations of the main experiment, the adversarial patch attack proceeds on the standard detector in the inference stage. To devise a defense-aware adaptive attack, we can regard the defensive frame-aware adaptive attack patch $p$ as the threatening patch $t$ in the above section. This is because we optimize the threatening patch $t$ with the knowledge of existence of shielding frame $s$, which is correlated with the defensive frame $d$ in the inference stage. Therefore, if the defense-aware adaptive adversarial patch $p$ added to input image, the new incoming input should be $x+p$. And the defended input should be $x''=x'+d=F_{0}(x+p)+d$. Here, $d$ is a universal defensive frame. This leads to the similar conclusion in Equation (\ref{eq:14}). Thus, for the adaptive attack, the gradient update for newly incoming adaptive attack ($x'$) is approximately the same with the gradient update of converged and balanced value in the final iteration $K_s$ due to the iterative competitive learning during the optimization process. Therefore, the experiment result in the manuscript demonstrates that our proposed method still shows the comparable adversarial robustness against adaptive patch attack.




\newpage


\ifCLASSOPTIONcaptionsoff
  \newpage
\fi



%
\bibliographystyle{IEEEtran}
\bibliography{IEEEabrv, refs}

\end{document}